\documentclass[10pt,twocolumn,letterpaper]{article}

\pdfoutput=1 

\usepackage{wacv}
\usepackage{times}
\usepackage{epsfig}
\usepackage{graphicx}
\usepackage{amsmath}
\usepackage{amssymb}
\usepackage{algorithm}
\usepackage{cite}
\usepackage[noend]{algpseudocode}
\usepackage[dvipsnames]{xcolor}
\usepackage{booktabs}
\usepackage{enumitem}
\setlist{nolistsep}
\usepackage{multirow}
\usepackage{subcaption}
\usepackage{authblk}

\usepackage[pagebackref=true,breaklinks=true,letterpaper=true,colorlinks,bookmarks=false,citecolor=blue,linkcolor=blue]{hyperref}

\wacvfinalcopy 

\def\wacvPaperID{41} 

\ifwacvfinal\fi
\begin{document}

\def\kh#1{\textcolor{blue}{{KH: }{#1}}}
\def\yht#1{\textcolor{red}{{YH: }{#1}}}
\def\BY#1{\textcolor{magenta}{{BY: }{#1}}}
\def\smaji#1{\textcolor{magenta}{{S: }{#1}}}
\def\jcsu#1{\textcolor{OliveGreen}{{JC: }{#1}}}
\def\new#1{{{#1}}}
\def\newnew#1{{{#1}}}
\def\mathbi#1{mkchandraker@eng.ucsd.edu\textbf{\em #1}}
\newcommand{\para }[1]{\medskip \noindent {\bf #1}}

\makeatletter
\renewcommand\AB@affilsepx{\qquad \protect\Affilfont}
\makeatother
\title{Active Adversarial Domain Adaptation}

\renewcommand*{\Authsep}{\qquad\qquad}
\renewcommand*{\Authand}{\qquad\qquad}
\renewcommand*{\Authands}{\qquad\qquad}
\author[1]{Jong-Chyi Su\thanks{Partial work done while at NEC Labs.}}
\author[2]{Yi-Hsuan Tsai}
\author[2]{Kihyuk Sohn\thanks{Currently at Google Cloud AI.}}
\author[2]{Buyu Liu}
\author[1]{Subhransu Maji}
\author[2,3]{Manmohan Chandraker}
\affil[1]{UMass Amherst}
\affil[2]{NEC Laboratories America}
\affil[3]{UC San Diego}
\maketitle
\thispagestyle{empty}

\begin{abstract}
We propose an active learning approach for transferring representations across domains.
Our approach, active adversarial domain adaptation (AADA), explores a duality between two related problems: \emph{adversarial domain alignment} and \emph{importance sampling} for adapting models across domains.
The former uses a domain discriminative model to align domains, while the latter utilizes the model to weigh samples to account for distribution shifts.
Specifically, our importance weight promotes unlabeled samples with large uncertainty in classification and diversity compared to labeled examples, thus serving as a sample selection scheme for active learning.
We show that these two views can be unified in one framework for domain adaptation and transfer learning when the source domain has many labeled examples while the target domain does not.
AADA provides significant improvements over fine-tuning based approaches and other sampling methods when the two domains are closely related.
Results on challenging domain adaptation tasks such as object detection demonstrate that the advantage over baseline approaches is retained even after hundreds of examples being actively annotated.

\end{abstract}


\vspace{-0.1in}
\section{Introduction}
\label{sec:intro}
The assumption that the training and test data are drawn from the same distribution may not be true in practical applications of machine learning and computer vision. Consequently, a predictor trained on the source domain ${\cal S}$ may perform poorly when evaluated on the target domain ${\cal T}$ different from the source.
This \textit{covariate shift} problem is common in many problems, \eg, the seasonal distribution of natural species may change in a camera trap dataset, or the image resolution can change from one dataset to another.

Many domain adaptation (DA) methods have been proposed to address this issue~\cite{daume2010frustratingly,ganin2016domain,uda-long2013,long2016unsupervised,tzeng2015simultaneous,tzeng2017adversarial,tzeng2014deep}.
The covariate shift assumes that the marginal distribution $p(x)$ of the data changes from ${\cal S}$ to ${\cal T}$, while the conditional label distribution $p(y|x)$ remains the same.
Domain adaptation methods operate by minimizing the differences of the marginal distributions of $x$ in the source domain {$p_{\cal S}(x)$} and target domain {$p_{\cal T}(x)$} by projecting the data through an embedding $\Phi(x)$, \eg, a deep network, while at the same time being predictive of the distribution $p_{\cal S}(y|x)$ in the source domain.
By matching the marginals, the covariate shift is reduced, thus improving the generalization of the model on the target domain compared to an ``unadapted'' model.

\begin{figure}[t]
\centering
    \includegraphics[clip, width=0.85\linewidth]{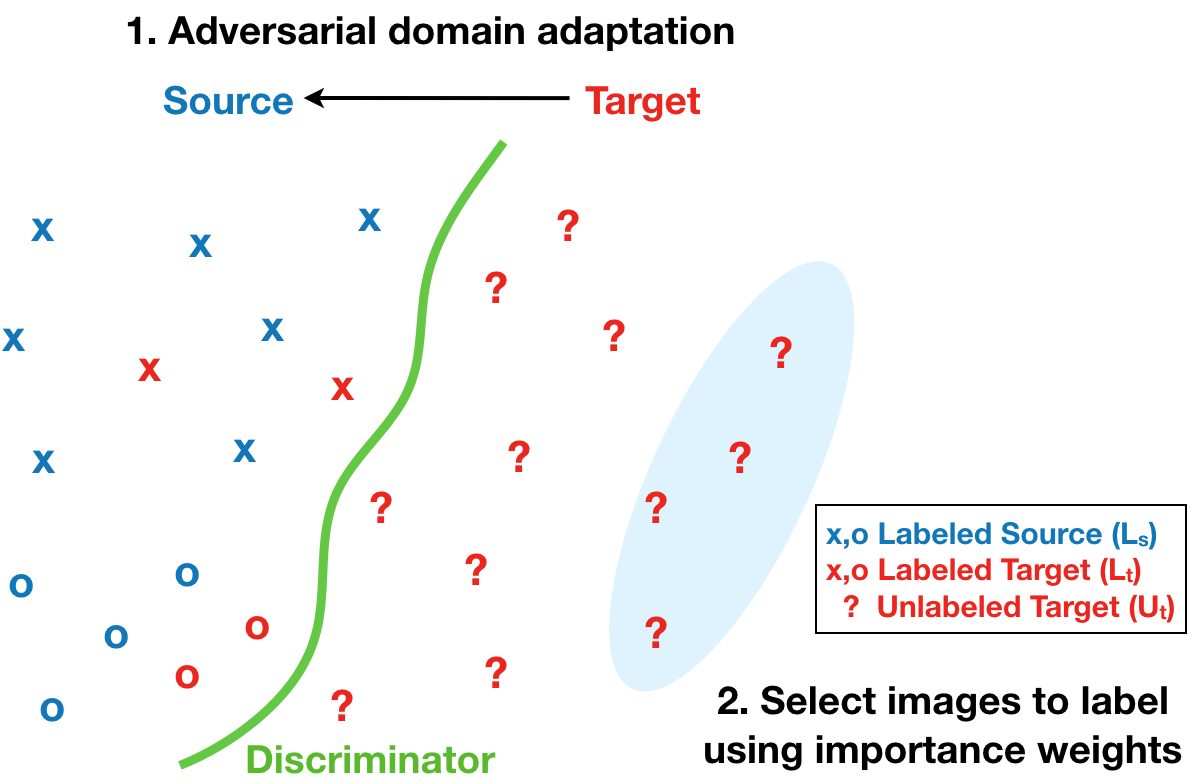}
\vspace{-0.1in}
\caption{
Source and target domain data are shown in blue and red. Circle and cross represent class labels, while question marks are unlabeled data.
We employ adversarial training to align features across the source and target domain, and use discriminator predictions to compute the \emph{importance weight} for sample selection of active learning.
}
\label{fig:teaser}
\vspace{-0.1in}
\end{figure}

While domain adaptation provides a good starting point, the performances of unsupervised DA methods often fall far behind their supervised counterparts~\cite{chen2018domain,Tsai_2018_CVPR}. In such cases, some labeled data from the target domain may bring in performance benefits. However, obtaining ground-truth annotations can be laborious and na\"ively collecting annotated data could be inefficient. In this work, we aim to answer the following questions: 1) how to select data to label from the target domain effectively, and 2) how to perform adaptation given these labeled data from the target domain.


To this end, we propose \emph{Active Adversarial Domain Adaptation} (AADA) that exploits the relation between domain adaptation and active learning to answer those questions.
%
Addressing our second question, we propose to adopt domain adversarial learning~\cite{ganin2016domain} between the union of labeled data from source/target and unlabeled target data, when the amount of labeled target data is small. However, after several rounds of active selection to accumulate many labeled data from the target domain, performing adversarial adaptation becomes counter-productive and simple transfer learning approaches (\eg, fine-tuning) serve the purpose.

Inspired by the importance weighted empirical risk minimization~\cite{sugiyama2007covariate,sugiyama2008direct}, we address our first question by proposing a sample selection criterion composed of the two cues: the \emph{diversity cue} and the \emph{uncertainty cue}. The diversity cue is from the \emph{importance} $w(x)\,{=}\, p_{\cal T}(x)/p_{\cal S}(x)$ where it can be estimated efficiently from the domain discriminator based on domain adversarial learning~\cite{goodfellow2014generative}. This allows one to sample unlabeled targets that are different from the labeled ones. The uncertainty cue is a lower bound to the empirical risk, which in our case is in the form of entropy of classification distribution. This promotes unlabeled data with low confidence for the next round of annotation.
The overall framework of our AADA is illustrated in Figure~\ref{fig:teaser}.

In experiments, we first validate the effectiveness of our approach on digit classification from SVHN to MNIST in Section~\ref{sec:ablation}, showing significant improvements over other baselines on domain adaptation, transfer learning, and active learning. Second, we conduct experiments for object recognition on the Office~\cite{office} and VisDA~\cite{peng2018visda} datasets with larger domain shifts in Section~\ref{sec:exp}. Last, we extend our method to object detection, adapting from the KITTI dataset~\cite{KITTI} to the Cityscapes dataset~\cite{Cityscapes}. The proposed AADA outperforms the fine-tuning baseline by 6\% when only 50 labeled images from the target domain are available. 

Finally, we summarize our contributions as follows:
\begin{itemize}
    \item An active learning framework by integrating domain adversarial learning and active learning for continuous semi-supervised domain adaptation.
    \item Improved classification performance with domain adversarial learning, while the discriminator prediction yields better importance weight for sampling.
    \item A connection between our sampling method and importance weight with domain adversarial training.  
    \item Reduced labeling cost on target domain on object classification and detection tasks.
\end{itemize}

\section{Related Work}
\subsection{Domain Adaptation}\label{sec:da}
Domain adaptation (DA) aims to make the model invariant to the data from the source and target domain.
%
For example, \cite{daume2010frustratingly} uses unlabeled data to measure the inconsistency between source and target domain classifiers.
%
%
Deep domain adaptation has been successful in recent years. The key idea is to measure the domain discrepancy at a certain layer of deep networks using domain discriminator~\cite{bousmalis2016domain,ganin2016domain} or maximum mean discrepancy (MMD) kernel~\cite{uda-long2013,long2016unsupervised,tzeng2015simultaneous,tzeng2014deep} and train CNNs to reduce the discrepancy.
Approaches that combine techniques from semi-supervised learning, such as entropy minimization~\cite{grandvalet2005semi,lee2013pseudo}, are proposed to enhance classification performance~\cite{long2016unsupervised,zhang2018collaborative}.
It has also been applied to more complicated vision tasks such as object detection~\cite{chen2018domain,inoue2018cross,Hsu_2019_CVPR_Workshops} and semantic segmentation~\cite{hoffman2017cycada,hoffman2016fcns,Tsai_2018_CVPR,Tsai_DA4Seg_ICCV19}, where the annotation cost is more expensive and how to select images to label become more crucial.

Different from the above-mentioned unsupervised DA, we explore the case where the budget is available to annotate a few labeled examples in the target domain.
Comparing to the method of~\cite{motiian2017few} which discusses how to train the model given few labeled targets with uniform distribution,
we focus on \emph{how to select target samples to label without knowing any prior distribution} of the target labels.

%
\subsection{Active Learning}
Active learning aims to maximize the performance with a limited annotation budget~\cite{cohn1995active,settles2012active}. 
Thus, the challenge is to quantify the $\textit{informativeness}$ of unlabeled data~\cite{kapoor2007active} so that they are maximally useful when annotated.
Many sampling strategies based on uncertainty~\cite{lewis1994heterogeneous,scheffer2001active}, diversity~\cite{dutt2016active,hoi2009semisupervised}, representativeness~\cite{xu2003representative}, reducing expected error~\cite{Roy2001,vijayanarasimhan2010visual} and maximizing expected label changes~\cite{freytag2014selecting,kading2015active,vezhnevets2012weakly} are studied and applied to vision tasks such as classification~\cite{qi2008two,joshi2009multi}, object detection~\cite{kao2018localization}, image segmentation~\cite{luo2013latent,sun2015active,vezhnevets2012weakly}, and human pose estimation~\cite{liu2017active}.
%
%
Among these, uncertainty sampling is simple and computationally efficient, making it a popular strategy in real-world applications.

Learning-based active learning methods~\cite{hsu2015active,konyushkova2017learning} are proposed recently by formulating a regression problem for the query procedure and learning strategies based on previous outcomes.
Deep active learning methods~\cite{sener2018active,shen2017deep} are studied for image classification and named-entity recognition.
\cite{mayer2018adversarial,zhu2017generative} propose to use generative models to synthesize data for training, but the performance is largely dependent on the quality of synthetic data, limiting their generality.
\begin{figure*}[t!]
\centering
    \includegraphics[clip, width=0.80\linewidth]{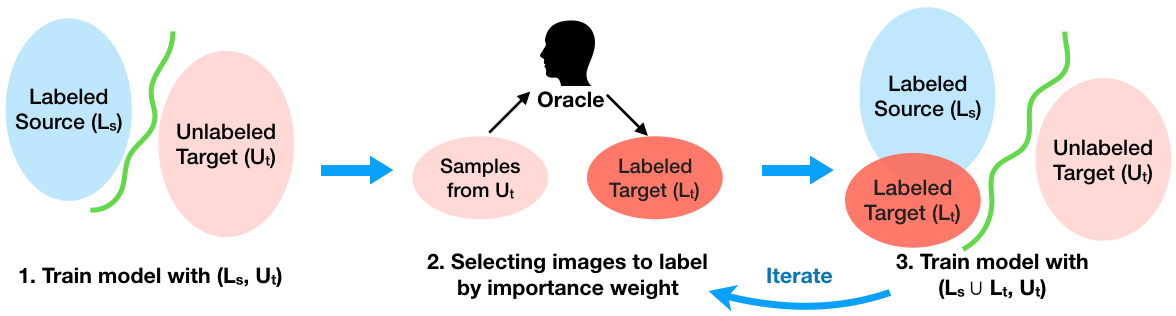}
\vspace{-0.1in}
\caption{Our proposed algorithm AADA. We start from an unsupervised domain adaptation setting with labeled source $L_s$ and unlabeled target $U_t$ data and train the model with domain adversarial loss. In each following round, we first select samples using importance weight from the unlabeled target domain to obtain annotations. We then re-train the model with labeled data $L_s \cup L_t$ and unlabeled data $U_t$.}
\label{fig:teaser2}
\vspace{-0.1in}
\end{figure*}

\subsection{Active Learning for Domain Adaptation}
%
Different from the aforementioned methods, we aim to unify active learning and domain adaptation.
Chattopadhyay~\etal~\cite{chattopadhyay2013joint} train the domain adaptation model with importance weights~\cite{bickel2009discriminative} and select samples by solving linear programming for minimizing the MMD distances between features. However, it is not clear how to incorporate this strategy with advanced techniques such as deep models and domain adversarial training.

The most relevant work is ALDA~\cite{rai2010domain,saha2011active}, which demonstrates its effectiveness in sentiment and landmine classification tasks.
ALDA trains three models, a source classifier $w_{src}$, a domain adaptive classifier $u_{\phi}$, and a domain separator $w_{ds}$. It first selects unlabeled target samples using $u_{\phi}$, and decide whether to acquire the label from $w_{src}$ (without cost) or the oracle (with cost) using $w_{ds}$. $u_{\phi}$ is then updated with the obtained labeled data. 
%

In addition to using the deep model, the proposed AADA is different from ALDA in several ways. First, our discriminator not only helps sample selection but also trains the recognition model adversarially to reduce the domain gap. Moreover, we combine diversity in the form of discriminator prediction and uncertainty in the form of entropy. To the best of our knowledge, we are the first to jointly tackle DA and active learning using neural networks on vision tasks.

\section{Proposed Algorithm}\label{sec:algorithm}
\new{In this section, we introduce our active adversarial domain adaptation (AADA). We begin with the background of domain adversarial neural networks in Section~\ref{sec:dann}, and then we motivate our sampling strategy by importance in Section~\ref{sec:sample}. The algorithm and its theoretical background under the semi-supervised domain adaptation setting are provided in Section~\ref{sec:aada_combined}.}
%
%
\subsection{Domain Adaptation}
\label{sec:dann}
In this section, we introduce the learning objective of our domain adaptation model. 
For simplicity, we describe the model in the image classification task. We denote $X$ as the input space and $Y\,{=}\,\{1,...,L\}$ as the label space. 
\new{The source data and (unlabeled) target data are drawn from the distribution $p_{\cal S}(x)$ and distribution $p_{\cal T}(x)$ respectively.}
%
We adopt the domain adversarial neural network (DANN)~\cite{ganin2016domain}, which is composed of three components: \emph{feature extractor} $G_f$ for the input $x$, \emph{class predictor} $G_y$ that predicts the class label $G_y(G_f(x))\,{\rightarrow}\,{\{1,...,L\}}$, and \emph{discriminator} $G_d$ that classifies the domain label $G_d(G_f(x))\,{\rightarrow}\,{\{0,1\}}$. We use $1$ for the source domain and $0$ for the target domain.
%
The objective function of the discriminator $G_d$ is defined as:
\begin{equation}
    \begin{split}
    \mathcal{L}_{d} & = \mathbb{E}_{x\sim{p_{\cal S}(x)}}\big[\log G_d(G_f(x))\big]\\
    & + \mathbb{E}_{x\sim{p_{\cal T}(x)}}\big[\log (1-G_d(G_f(x)))\big],
    \end{split}
    \label{eq:discr}
\end{equation}
where $G_f,G_y,G_d$ are parameterized by $\theta_f,\theta_y,\theta_d$, respectively.
To perform domain alignment, features generated from $G_f$ should be able to fool the discriminator $G_d$, 
and hence we adopt an adversarial loss to form a min-max game:
\begin{equation}
    \min_{\substack{\theta_f,\theta_y}} \max_{\substack{\theta_{d}}} \mathcal{L}_c(G_y(G_{f}(x)),y) + \lambda\mathcal{L}_d,
    \label{eq:adv_loss}
\end{equation}
where $\mathcal{L}_c$ is the cross-entropy loss for classification, $y$ is the class label, and $\lambda$ is the weight between two losses. 
%
%
%
%

\subsection{Sample Selection}\label{sec:sample}
Given an unsupervised domain adaptation setting where labeled data is only available from the source domain, the goal of our sample selection is to find the most informative data from the unlabeled target domain.
We motivate the sample selection criteria from the idea of importance weighted empirical risk minimization (IWERM)~\cite{sugiyama2007covariate}, whose learning objective is defined as follows:
\begin{equation}
\min_{\theta_{f}, \theta_{y}}\mathbb{E}_{(x,y)\sim {p_{\cal S}(x,y)}}\Big[\,\frac{p_{\cal T}(x)}{p_{\cal S}(x)}\mathcal{L}_{c}\big(G_{y}(G_{f}(x)), y\big)\Big],
\end{equation}
where $w(x)\,{=}\,\frac{p_{\cal T}(x)}{p_{\cal S}(x)}$ is an importance of each labeled data in the source domain. The formulation indicates which data is more important during optimization. First, the data with higher empirical risk $\mathcal{L}_{c}\big(G_{y}(G_{f}(x)), y\big)$, and second, the one with higher importance, \ie, larger density in the target distribution $p_{\cal T}(x)$ but lower in the source $p_{\cal S}(x)$.

Unfortunately, applying this intuition to come up with a sample selection strategy is non-trivial. This is because the target data is mostly unlabeled and the empirical risk cannot be computed before annotation. Another problem is that the importance estimation of high-dimensional data is difficult~\cite{sugiyama2008direct}.
We take advantage of domain discriminator to resolve the second issue. Note that, with adversarial training, the optimal discriminator~\cite{goodfellow2014generative} is obtained at 
\begin{equation}
G_d^*(\hat{x})\,{=}\,\frac{p_{\cal S}(x)}{p_{\cal S}(x)\,{+}\,p_{\cal T}(x)}\Rightarrow w(x)\,{=}\,\frac{1\,{-}\,G^*_d(\hat{x})}{G^*_d(\hat{x})},
\end{equation}
where $\hat{x}\,{=}\,G_{f}(x)$. Next, assuming cross-entropy as an empirical risk, we resolve the first issue by measuring the entropy of unlabeled data, which is a lower bound to the cross-entropy.\footnote{$H(p,q)\,{=}\,D_{KL}(p||q)\,{+}\,H(p)\, {\geq}\, H(p)$.}
Finally, our sample selection criterion $s(x)$ for unlabeled target data is written as follows:
\begin{equation}\label{eq:importance_weight}
s(x)=\frac{1\,{-}\,G^*_d(G_{f}(x))}{G^*_d(G_{f}(x))}\mathcal{H}(G_y(G_{f}(x))).
\end{equation}
Two components in the measure are interpreted as follows: 1) \emph{diversity} cue $(1{-}G^*_d(G_{f}(x)))/G^*_d(G_{f}(x))$, and 2) \emph{uncertainty} cue $\mathcal{H}(G_y(G_{f}(x)))$. 
The diversity cue allows us to select unlabeled target data which is less similar to the labeled ones in the source domain, while the uncertainty cue suggests data that the model cannot predict confidently.

\subsection{Active Adversarial Domain Adaptation}
\label{sec:aada_combined}
Based on the two objectives of domain adaptation and sample selection, we explain the role of these two components in their collaboration for active learning for domain adaptation purposes.
\begin{algorithm}[!t]
\caption{AADA}\label{alg:AADA}
\begin{algorithmic}
\State \textbf{Input:} labeled source $L_s$; unlabeled target $U_t$; \\
\quad\quad~~~ labeled target $L_t=\emptyset$; budget per round $b$
\State \textbf{Model:} $\cal{M}$=\{$G_f$, $G_y$, $G_d$\}; feature extractor $G_f$; \\
\quad\quad~~~ class predictor $G_y$; discriminator $G_d$
\State Train $\cal{M}$ with $(L_s, U_t)$
\For{round $\gets$ 1 to MaxRound}
\State Compute $s(x) ~\forall x\in U_t$ via~\eqref{eq:importance_weight}
\State Select a set of $b$ images $z$ from $U_t$ according to $s(z)$
\State Get labels $y_z$ from oracle
\State $L_t \leftarrow L_t \cup (z,y_z) $
\State $U_t \leftarrow U_t \setminus (z,y_z) $
\State Train $\cal{M}$ with $(L_s\cup L_t,U_t)$
\EndFor
\end{algorithmic}
\end{algorithm}

\para{Collaborative Roles.}
For domain adaptation, the goal is to learn domain-invariant features via \eqref{eq:adv_loss} that better serves as a starting point for the next sample selection step.
During the adversarial learning process, a discriminator is learned to separate source and target data, and thus we can utilize its output prediction as an indication for selection via the importance weight in \eqref{eq:importance_weight}.
By iteratively performing adversarial learning and active learning, the proposed method gradually selects informative samples for annotations guided by the domain discriminator, and then these selected samples are used for supervised training to minimize the domain gap, in a collaborative manner.

One may still obtain a discriminator without adversarial learning and it can be easily learned to separate samples across two different domains. \newnew{However, learning a discriminator in this way can be problematic for active learning. First, this discriminator may give identically high scores to most target samples. Thus it lacks the capability of selecting informative ones. Moreover, the learned classifier and this discriminator may focus on different properties if they are not learned jointly. If this is the case, the informative samples that current discriminator selects are not necessarily beneficial for classifier update.}
%
We provide more evidence for the necessity of adversarial training in Section~\ref{sec:training}.

\para{Active Learning Process.}
Our overall active learning framework is illustrated in Figure~\ref{fig:teaser2}.
%
We start our AADA algorithm by learning a DANN model in an unsupervised domain adaptation setting as described in Section~\ref{sec:dann}, and then use the learned discriminator to perform the initial round of sample selection from all unlabeled target samples based on~\eqref{eq:importance_weight}. Once obtaining the selected samples, we acquire their ground-truth labels.
%

%
For the following rounds, 
we have a small set of labeled target data $L_t\,{\sim}\,{p_{\cal T}(x,y)}$, a set of labeled source data $L_s\,{\sim}\,{p_{\cal S}(x,y)}$, and the remaining unlabeled target data $U_t\,{\sim}\,{p_{\cal T}(x)}$.
Thus, the learning setting is different from the initial stage as we now have labeled domains $L_s$ and $L_t$. To accommodate labeled data from both domains, we revisit an analysis of domain adaptation~\cite{ben2010theory,blitzer2008learning} whose generalization bound is given as:
\begin{align}
\epsilon_{\cal T}(\hat{h}) &\leq {\epsilon}_{\cal T}(h_{\cal T}^{*}) + \gamma_{\alpha} +  d_{\mathcal{H}\Delta\mathcal{H}}(L_{s}\,{\cup}\,L_{t},U_{t})\label{eq:dann-theory}\\
&+ 4\sqrt{\left(\frac{\alpha_{s}^2}{\beta_{s}}+\frac{\alpha_{t}^2}{\beta_{t}}\right)\left(\frac{d\log(2m)-\log(\delta)}{2m}\right)},\nonumber
\end{align}
with $\gamma_{\alpha}\,{=}\,\epsilon_{\cal T}(h)\,{+}\,\alpha_{s}\epsilon_{\cal S}(h)\,{+}\,\alpha_{t}\epsilon_{\cal T}(h)$, $m$ is the number of labeled examples, $d$ is VC-dimension of hypothesis class and $h$ is the hypothesis (\ie, classifier). $\alpha{=}(\alpha_{s},\alpha_{t})$ is a weight vector between the errors of labeled source and labeled target, while $\beta{=}(\beta_{s},\beta_{t})$ is a proportion of labeled examples for source and target domains. Assuming zero error on the labeled examples (\ie, $\epsilon_{\cal S}(h)\,{=}\,\epsilon_{\cal T}(h)\,{=}\,0$), the bound is the tightest when $\alpha_{s}\,{=}\,\beta_{s}$ and $\alpha_{t}\,{=}\,\beta_{t}$. 

%
This leads us training a new model that adapts from all labeled data $L_s\,{\cup}\,L_t$ to unlabeled data $U_t$ with uniform sampling of individual examples from labeled set to ensure the tightest bound. Thus, we use uniform sampling of labeled source and target examples for sampling batches during training unless otherwise stated. 
%
Then, we select candidates from \newnew{the remaining unlabeled target set} $U_t$ based on the new discriminator $G_d$ and new classifier $G_y$ following the same importance sampling strategy for the next round of training.
%
The overall algorithm is shown in Algorithm~\ref{alg:AADA}. 
%


\section{Experiments on Digit Classification}\label{sec:ablation}

\begin{figure*}[th!]
    \newcommand\wi{0.47\linewidth}
    \newcommand\wifig{0.9\textwidth}
    \centering
    \begin{subfigure}{\wi}
        \centering 
        \includegraphics[clip, width=\wifig]{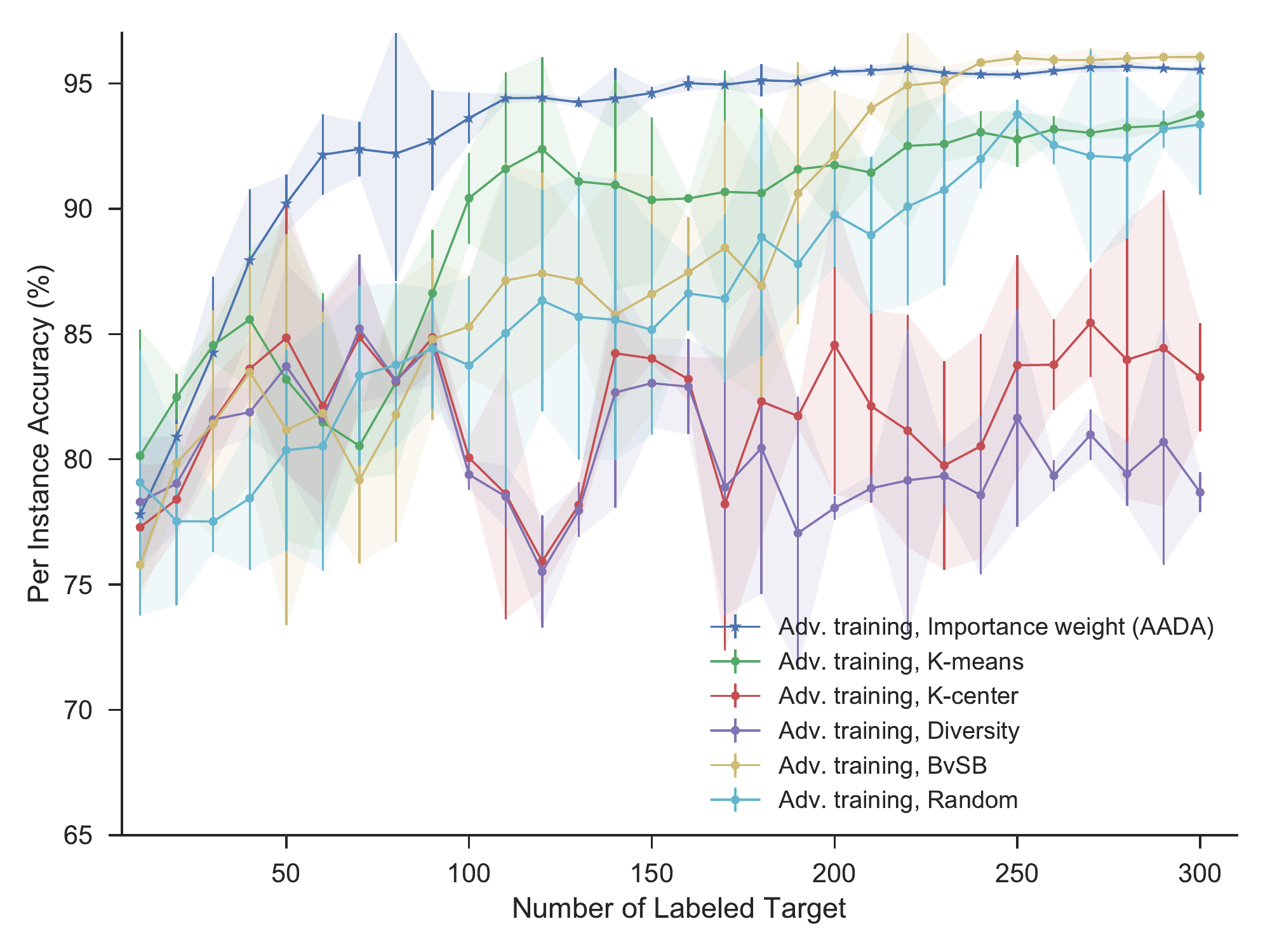}
        \vspace{-0.1in}
        \caption{\new{Different sampling strategies with adversarial training.}}
        \label{fig:ab_adv}
    \end{subfigure}
    \quad
    \begin{subfigure}{\wi}
        \centering 
        \includegraphics[clip, width=\wifig]{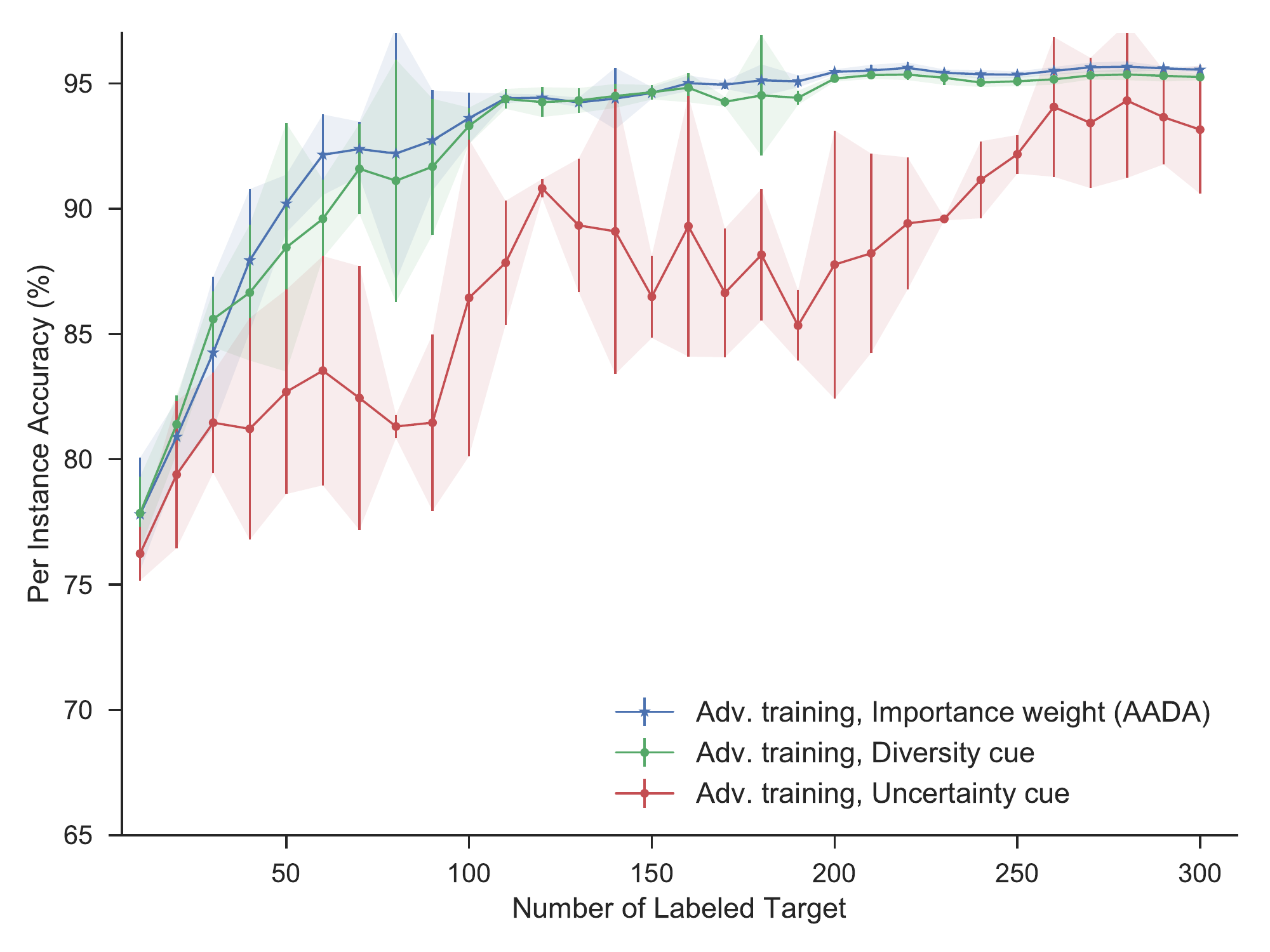}
        \vspace{-0.1in}
        \caption{\new{Different sampling cues with adversarial training.}}
        \label{fig:ab_cues}
    \end{subfigure}
    \vspace{-0.02in} \\
    \begin{subfigure}{\wi}
        \centering
        \includegraphics[clip, width=\wifig]{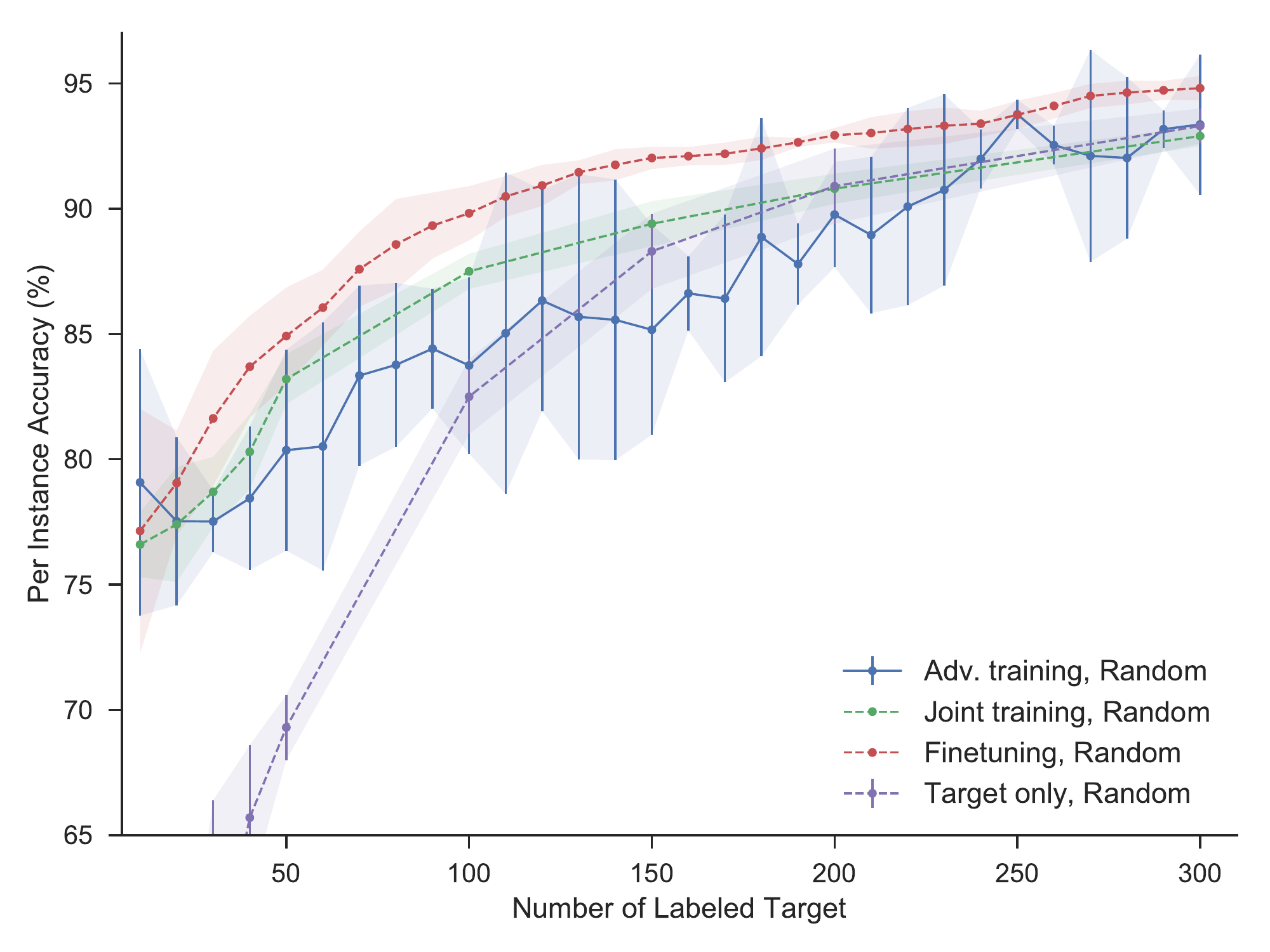}
        \vspace{-0.1in}
        \caption{\new{Different training schemes with random sampling.}}
        \label{fig:ab_random}
    \end{subfigure}
    \quad
    \begin{subfigure}{\wi}
        \centering 
        \includegraphics[clip, width=\wifig]{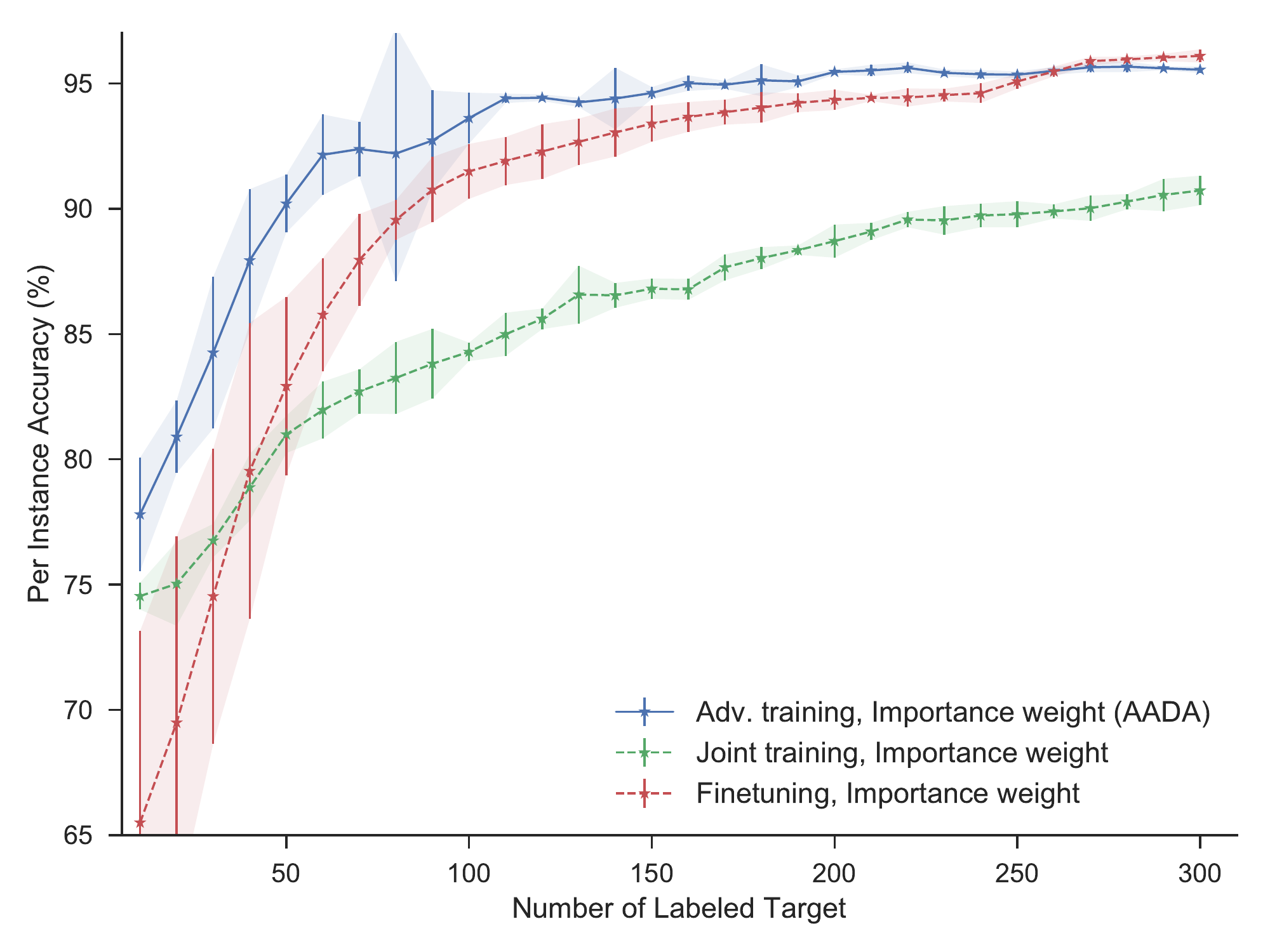}
        \vspace{-0.1in}
        \caption{\new{Different training schemes with importance weight.} 
        \label{fig:ab_imp}
        }
    \end{subfigure}
    \vspace{-0.1in} 
    \caption{\new{Ablation studies on digit classification (SVHN $\rightarrow$ MNIST). Each data point is the mean accuracy over five runs, and the error bar shows the standard deviation. We show that: (a) sampling using importance weight performs the best when using adversarial training, (b) combining diversity and uncertainty cues performs better for selecting samples, (c) fine-tuning is the best training scheme when random sampling is used, (d) when using importance weight for sampling, adversarial training is the best when there are less than 250 labeled target. Overall, our AADA which uses adversarial training and importance weigh provides the best performance when few labeled targets are available.}}
    \label{fig:ablation}
    \vspace{-0.1in} 
\end{figure*}

As discussed above, our proposed method aims to address two questions: 1) how to select images to label from $U_t$ to yield the most performance gain? and 2) how to train a classifier given $\{L_s,L_t,U_t\}$? Our experiments then consists of our explorations for both components. In this section, we first perform detailed experiments in a mix-and-match way on the digit classification task from SVHN~\cite{svhn} to MNIST~\cite{mnist}. Specifically,
we explore the following {\bf{training schemes}}:
{\flushleft {\bf 1) Adversarial Training:}} we train the classifier via~\eqref{eq:adv_loss} using $(L_s\cup L_t, U_t)$.
{\flushleft {\bf 2) Joint Training:}} we train the classifier in a supervised way using $L_s\cup L_t$. Note that we still train a discriminator for sample selection but without adversarial training.
{\flushleft {\bf 3) Fine-tuning:}} we train a classifier using $L_s$ and then fine-tune it on $L_t$, both in a supervised way. Discriminator is trained in a similar manner to Joint Training.
{\flushleft {\bf 4) Target Only:}} we train our classifier with $L_t$ only.
\vspace{0.08in}
\\
The {\bf{sampling strategies}} we explored are:
\vspace{-0.08in}
{\flushleft {\bf 1) Importance Weight:}} we select samples based on the proposed importance weight $s(x)$~\eqref{eq:importance_weight}.
{\flushleft {\bf 2) K-means Clustering:}} we perform k-means clustering on image features $G_f(x), \forall x \in U_t$, where the number of clusters is set to $b$ in each round. For each cluster, we select one sample which is the closest to its center.
{\flushleft {\bf 3) K-center (Core-set)~\cite{sener2018active}:}} \newnew{we use greedy k-center clustering to select b images $z$ from $U_t$ such that the largest distance between unlabeled data $U_t \setminus z$ and labeled data $L_t \cup z$ is minimized. We use L2 distance between image features $G_f(x)$ for the measurement.}
{\flushleft {\bf 4) Diversity~\cite{dutt2016active}:}} for each unlabeled sample in $U_t$, we compute its distance to all samples 
in $L_t$ and obtain the average distance. Then we rank unlabeled samples w.r.t.~its average distance in descending order and select the top $b$ samples. L2 distance is applied on features $G_f(x)$.
{\flushleft {\bf 5) Best-versus-Second Best (BvSB)~\cite{joshi2009multi}:}} we use the difference between the highest and the second highest class prediction as the uncertainty measure., \ie, $\max_i G_{y_i}(\hat{x}) - G_{y_j}(\hat{x})$, where class $j$ has the second highest prediction.
{\flushleft {\bf 6) Random Selection:}} we select samples uniformly at random from all the unlabeled target data $U_t$.

\vspace{0.1in}

Our AADA uses importance weight for sample selection, and adversarial training as the training scheme.
We note that other unsupervised DA methods can be orthogonal to our approach, \eg, one can use improved DANN such as CyCADA~\cite{hoffman2017cycada} for initialization but still use our criteria for selecting samples to label. Here we focus on sample selection and only use the vanilla adversarial training.
We also note that different sampling methods do not compete with AADA as they can be combined with our method. For example, BvSB can be used as an alternative uncertainty measurement as opposed to entropy in~\eqref{eq:importance_weight}.

\vspace{-0.1in}
\paragraph{Experimental Setting.} Commonly in the active learning literature~\cite{luo2013latent,sun2015active}, we simulate oracle annotations by using the ground-truth in all our experiments. 
We consider an adaptation task from SVHN to MNIST, where the former and latter are initially considered as labeled source ${L}_s$ and unlabeled target ${U}_t$ respectively. SVHN contains 73,257 RGB images and MNIST consists of 60,000 grayscale images, both from the digit classes of $0$ to $9$. Not only differ in color statistics, the images from two datasets also experience different local deformations, making the adaptation task challenging.
For this task, we use the variant of LeNet architecture~\cite{hoffman2017cycada} and add an entropy minimization loss ${\mathcal L}_{ent}={\mathcal H}(G_y(G_f(x))$ for regularization~\cite{long2016unsupervised} during training. For each round, we train the model for 60 epochs using Adam~\cite{kingma2014adam} optimizer with learning rate $\{2\times10^{-4},1\times10^{-4},5\times10^{-5}\}$ for 20 epochs each. The batch size is 128 and $\lambda=0.1$. We set budget to 10 in each round and perform 30 rounds, eventually selecting 300 images in total from the target domain.
We carry our experiments with five different random seeds and report the averaged accuracy after each round. 
We use PyTorch~\cite{pytorch} for our implementation.
\new{
\subsection{Comparison of Sampling Methods}\label{sec:sampling}
We start from comparing different sampling method combined with adversarial training.
As shown in Figure~\ref{fig:ab_adv}, importance weight often outperforms its active sampling counterparts. It can achieve $95\%$ accuracy with 160 samples after 16 rounds while the random selection baseline requires two times more annotations to have similar performance. 
Moreover, our proposed method consistently improves performance when more samples are selected and annotated, whereas other baselines generate unstable performances.
One reason for such observation is that the class distribution of the selected samples in each round is not uniform. If the selected targets are heavily biased towards few classes, the ``mode collapse'' issue due to adversarial training gives high test accuracy on those classes but low accuracy on others, causing the overall lower accuracy. However, sampling with importance weight makes the result more stable after each round.
As a reference, AADA performs similarly as random selection ($97.5\%$ accuracy) with 1000 labeled targets. The performance saturates at around $99.0\%$ accuracy with 5000 labeled targets and achieves $99.5\%$ accuracy with all 73,257 labeled targets.
}
\vspace{-0.1in} 
\new{
\subsection{Comparison of Different Cues}\label{sec:cues}
We perform an ablation study of the two components in the proposed importance weight~\eqref{eq:importance_weight}. The \emph{diversity} cue, \ie, $\frac{1\,{-}\,G^*_d(G_{f}(x))}{G^*_d(G_{f}(x))}$, uses the predictions from the discriminator $G_d$, while the \emph{uncertainty} cue $\mathcal{H}(G_y(G_{f}(x)))$ uses the predictions from the classifier $G_y$. As shown in Figure~\ref{fig:ab_cues}, using diversity cue outperforms that of uncertainty cue, while combining these two yields the best performance.
However, the benefits of using different cues may depend on the characteristics of each dataset and will be discussed later.
}
\new{
\subsection{Comparison of Training Schemes}\label{sec:training}
We compare different training schemes and show the effectiveness of combining adversarial training with importance weight.
First, we provide a study of four training schemes in Figure~\ref{fig:ab_random}, all using random sampling. In this case, we find that adversarial training suffers from mode collapse problem and fine-tuning is the best option. Fine-tuning is also the most effective and widely-used method of transfer learning as discovered in the deep learning literature~\cite{sharif2014cnn,yosinski2014transferable}.
}

\new{
However, once the imbalance sampling problem can be effectively addressed, \eg using the proposed importance weight, we can benefit from adversarial training. Figure~\ref{fig:ab_imp} demonstrates the effectiveness of combining adversarial training with importance weight. We can see that it outperforms all the settings in Figure~\ref{fig:ab_random}.
Moreover, our AADA method demonstrates its effectiveness especially when very few labeled targets $L_t$ are available; on the other hand, when more and more labeled targets are available, fine-tuning seems to be a better option as the benefit of leveraging information from source domain has decreased (as explained in Section~\ref{sec:intro}). In our experiment, using fine-tuning performs better than using adversarial training when there are more than 250 labeled target selected using importance weight.
%
}
\vspace{-0.1in}
\new{
\paragraph{Comparison with ALDA~\cite{saha2011active}.}
\newnew{For the baseline of using joint training and importance weight, we train the classifier $G_y$ and the feature extractor $G_f$ with $L_s \cup L_t$, and train the discriminator $G_d$ for separating labeled and unlabeled data. The two objectives are trained jointly but not adversarially.} This can be seen as an extension of ALDA~\cite{saha2011active} using deep learning framework, despite some differences such as 1) the use of joint training instead of updating a perceptron, and 2) selecting samples using our proposed importance weight instead of using the margins to the linear classifier.
}

\new{
Interestingly, this baseline (as shown in Figure~\ref{fig:ab_imp}) is worse than the one using joint training and random sampling (as shown in Figure~\ref{fig:ab_random}). 
This is mainly due to the lack of diversity. Specifically, without the help of adversarial loss, the importance weight can be very confident thus lacks the ability to provide sufficient diverse samples. This problem also remains for the original ALDA~\cite{saha2011active} method. Again, as shown in Figure~\ref{fig:ab_imp}, our AADA outperforms this baseline by $7.6\%$ on average of the first 25 rounds, showing that adversarial training not only helps adapt the model but also collaborates with importance weight for sampling. 
}
%

\section{More Experimental Results}\label{sec:exp}
\new{In this section, we conduct experiments on object recognition and object detection datasets. Here we focus on comparing different sampling methods and refer the readers to supplementary material for complete comparisons.} 



\subsection{Object Recognition}\label{sec:office}
We validate our idea on the Office domain adaptation dataset~\cite{office}. It consists of 31 classes and three domains: amazon (A), webcam (W), and dslr (D), each with $\{2817, 795, 498\}$ images.
Specifically, we select dslr (D) as the source domain and amazon (A) as the target one. 
We further split the target domain using the first $2/3$ images as ${U}_t$ and the rest as the test set to evaluate all methods. 
We utilize ResNet-18~\cite{resnet} model (before the first \texttt{fc} layer) pre-trained on ImageNet as the feature extractor $G_f$. On top of it, $G_y$ has one layer while $G_d$ has \texttt{fc-ReLU-fc} with 256-256-2 channels.
We train our model with SGD for 30 epochs with a learning rate of $0.005$. The batch size is 32 and $\lambda=0.0001$. Budget per-round $b$ is set to 50 and we perform 20 rounds in total. We start the first round with random selection for all the methods as a warm-up.

\begin{figure}[t!]
\centering
    \includegraphics[clip, width=0.90\linewidth]{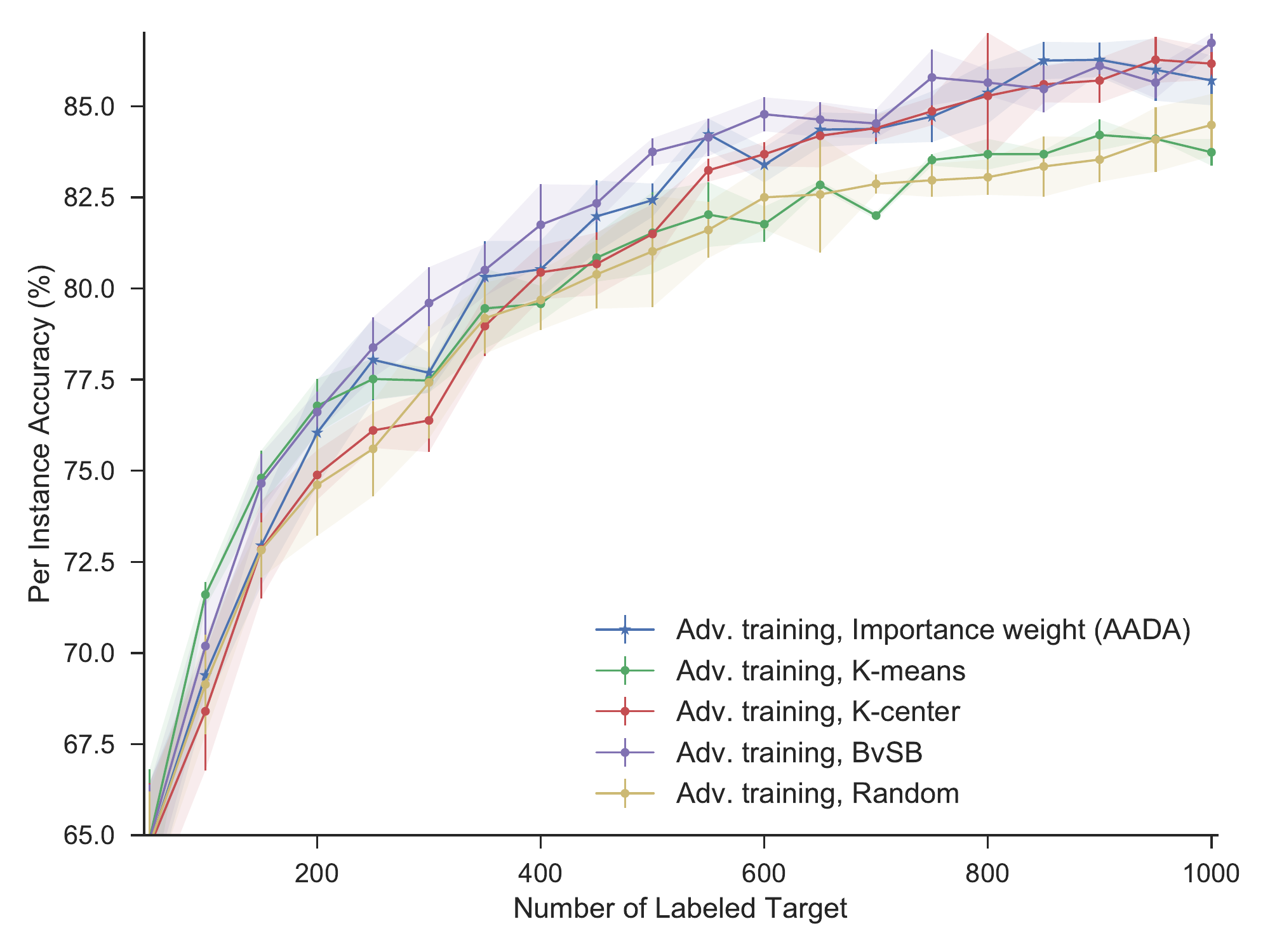}
\vspace{-0.1in}
\caption{Object classification result (Office D $\rightarrow$ A). \new{We compare different sampling methods with adversarial training. BvSB and AADA perform the best with 81.3\% and 80.7\% mean accuracy of 20 rounds separately.}
}
\label{fig:office}
\vspace{-0.1in}
\end{figure}

Figure~\ref{fig:office} demonstrates different sampling baselines with adversarial training. Our AADA method performs competitively with BvSB and outperforms all other methods, suggesting that the uncertainty cue is more useful in this dataset. More specifically, AADA outperforms random selection by around $3\%$ from round 10 to round 20, and our AADA can achieve $85\%$ accuracy with 800 labeled targets while random selection requires 200 more to achieve similar performance. \new{Note that BvSB is one of the variants of our method, which also deploys our adversarial training scheme and uncertainty measurement.}

\subsection{Object Detection}\label{sec:detection}
Now we focus on object detection task adapting from KITTI~\cite{KITTI} to Cityscapes~\cite{Cityscapes}.
We use the same setting as~\cite{chen2018domain}, which only considers the car object and resizes images to 500 for the shorter edge while keeping the aspect ratio. After discarding images without cars, we obtain 6,221 and 2,824 training images from KITTI and Cityscapes respectively, and we split 500 images from Cityscapes for testing. 
Mean average precision at 0.5 IoU (mAP@0.5) is our evaluation metric in this task~\cite{chen2018domain,inoue2018cross}. 
We adopt Faster-RCNN~\cite{Ren_etal_2015} with the ResNet-50 architecture combining with FPN~\cite{Lin_etal_2017b} as the feature extractor, and perform image-level adaptation as proposed in~\cite{chen2018domain}. 
We select $\{10, 10, 10, 20, 50, 100\}$ images in each round and assume that the cost of labeling one image is the same.

\begin{figure}[t!]
\centering
    \includegraphics[clip, width=0.99\linewidth]{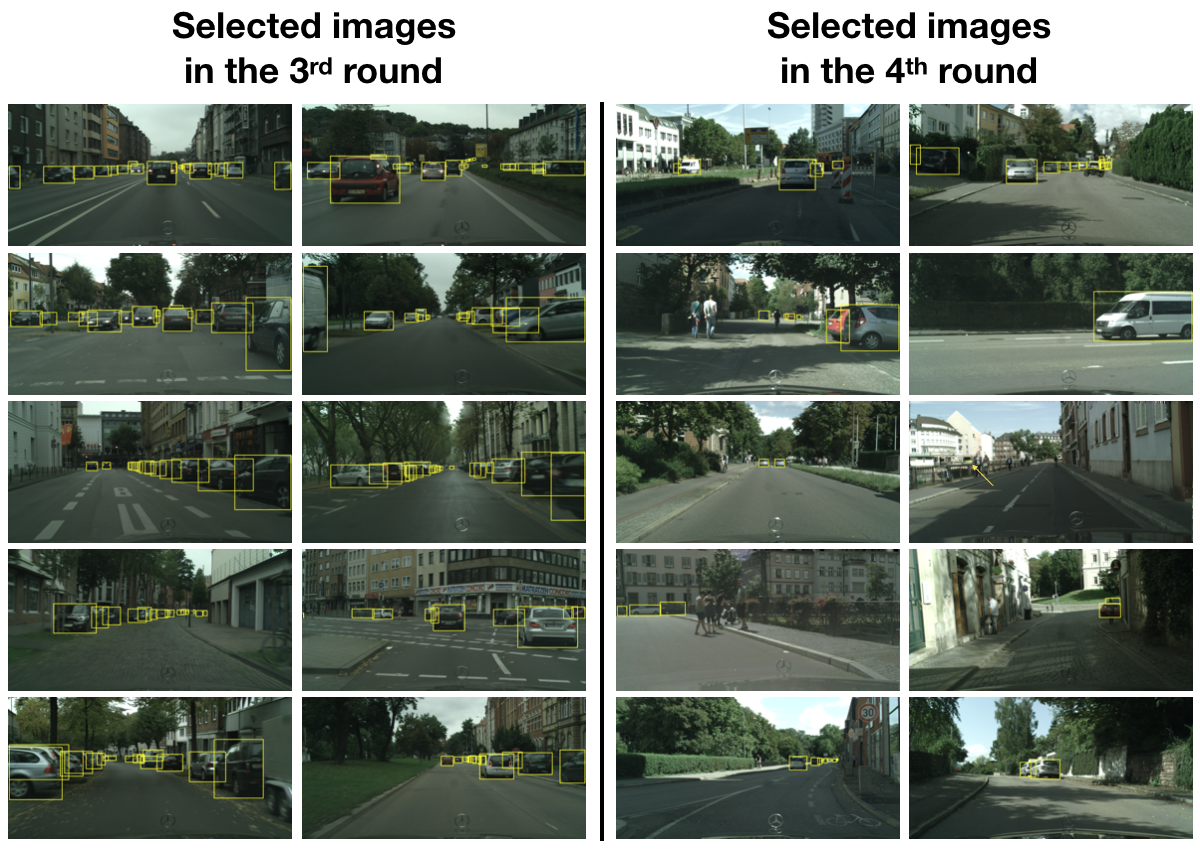}
\vspace{-0.1in}
\caption{
\new{Top 10 images selected in the third and the fourth rounds from the target domain (Cityscapes) using AADA. The ground-truth bounding boxes of cars are shown in yellow. Images selected in the third round have more cars \newnew{and the semantic layouts are different} w.r.t.~that of the fourth round, showing that diverse samples are selected by AADA.}}
\label{fig:visualization_det}
\vspace{-0.in}
\end{figure}

\new{We report our quantitative results in Table~\ref{tab:detection}. Our baselines 
include adversarial training with other sampling methods and different training schemes with random sampling. 
Note that BvSB is not included here due to the fact that in the single object category detection scenario, it provides similar measurement as entropy.
Overall, using adversarial training and importance weight (AADA) gives the best performance. 
Specifically, $60.4\%$ accuracy can be achieved with 100 labeled target selected by AADA, while other baselines require about twice as much annotations to achieve similar performance. 
We further illustrate images selected with AADA within two rounds in Figure~\ref{fig:visualization_det}. As can be seen in this figure, we are able to select diverse images with different semantic layouts}.


\begin{table}[t!]
\footnotesize{
	\renewcommand{\arraystretch}{1}
    \setlength{\tabcolsep}{5.5pt}
    \centering
	\begin{tabular}{@{\hskip3pt}c@{\hskip3pt}|@{\hskip3pt}c@{\hskip3pt}|c|c|c|c|c|c}
		\toprule
		\multirow{2}{*}{Training} & \multirow{2}{*}{Sampling} & \multicolumn{6}{c}{Number of Labeled Target} \\
		\cline{3-8}
		 & & 10 & 20 & 30 & 50 & 100 & 200\\
		\hline
		Adversarial & Imp.~weight & \textbf{49.4} & \textbf{53.3} & \textbf{54.6} & \textbf{57.4} & \textbf{60.4} & \textbf{62.3}\\
		Adversarial & K-means & 49.1 & 51.7 & 53.8 & 56.8 & 59.2 & 60.9\\
		Adversarial & Entropy & 48.9&50.9&52.3&54.3&58.1&61.0\\
		Adversarial & Random & 47.4 & 49.8 & 51.6 & 55.2 & 58.6 & 61.7\\
		Joint & Imp.~weight & 48.5 & 52.1 & 53.5 & 56.2 & 58.6 & 60.5\\
		Joint & Random & 45.5 & 48.8 & 51.8 & 54.9 & 59.0 & 61.6\\
		Fine-tuning & Random & 41.0 & 46.0 & 48.7 & 51.4 & 56.0 &  59.8\\
		Target only & Random & 29.0 & 38.5 & 42.1 & 48.3 & 53.3 & 58.8\\
		\bottomrule
	\end{tabular}
	\vspace{-0.1in}
	\caption{
	\new{Object detection results (KITTI $\rightarrow$ Cityscapes). Our AADA method (first row) outperforms all other baselines, including using adversarial training and other sample selection methods, as well as using different training schemes and random sampling.}}
	\label{tab:detection}
	\vspace{-0.1in}
}
\end{table}


\subsection{VisDA-18 Challenge}\label{sec:visda}


We investigate the VisDA-18 domain adaptation challenge~\cite{peng2018visda,peng2018syn2real} as a special case. The source domain is composed of 78,222 synthetic images across 12 object categories rendered from 3D CAD models, while the target domain contains 5,534 real images. 
We consider the 12-way classification problem following the setting in~\cite{peng2018syn2real} and the ImageNet pre-trained ResNet-18~\cite{resnet} model is used as a feature extractor. As mentioned in~\cite{peng2018syn2real}, without using ImageNet pre-training, the accuracy would be very low and unsupervised domain adaption methods do not work. 
However, ImageNet images are closer to the target domain, this raises our interest to investigate whether images from source domain still help in this scenario. 
%
\begin{figure}[t!]
\centering
    \includegraphics[clip, width=0.90\linewidth]{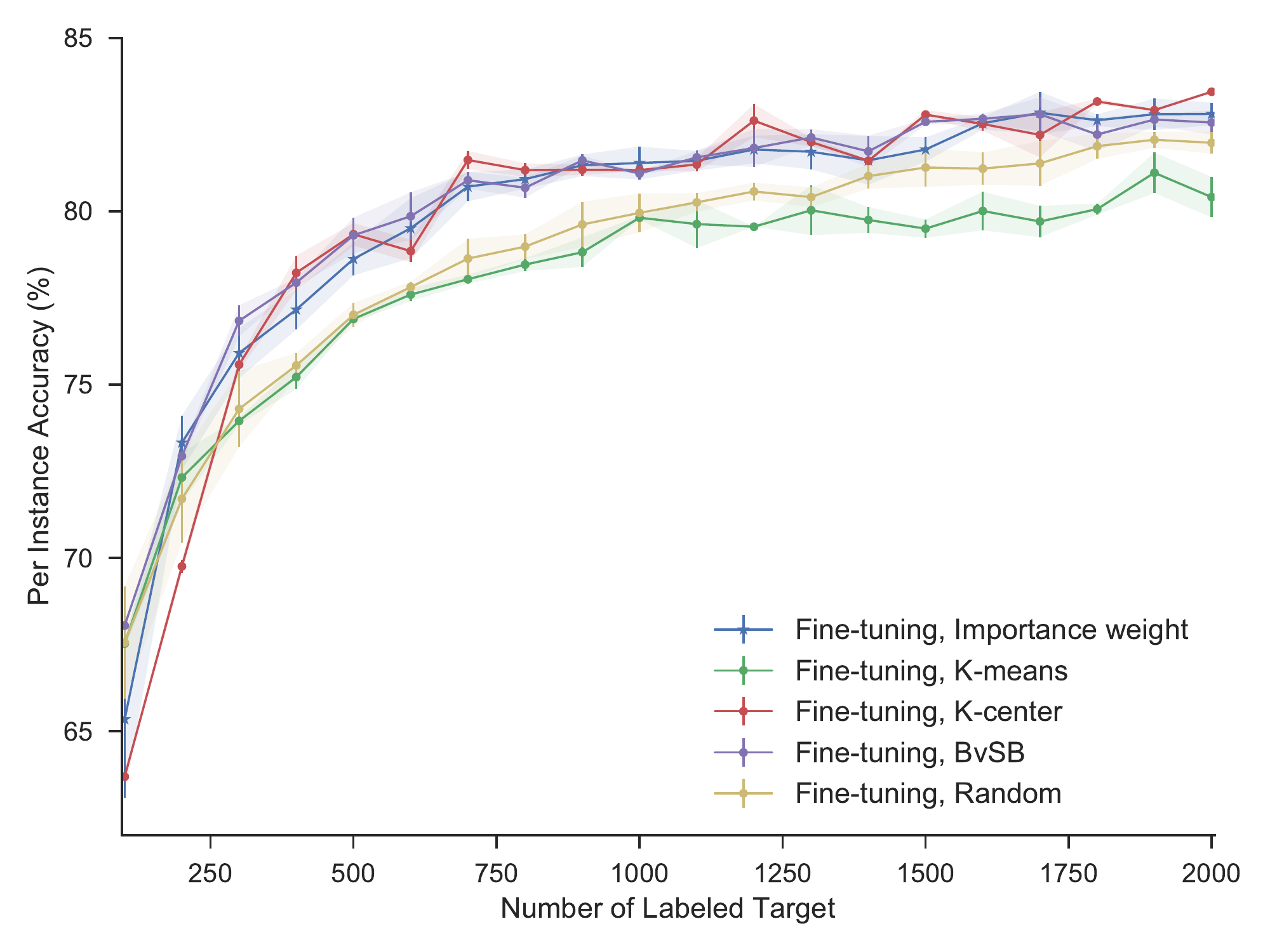}
\vspace{-0.1in}
\caption{VisDA-18 result (synthetic $\rightarrow$ real). Here we use fine-tuning as the training scheme and compare different sampling strategies. \new{Using importance weight for sampling performs equally well as BvSB and k-center baselines, and outperforms k-means and random baselines. The mean accuracies after each round are 79.8\% and 80.1\% for importance weight and BvSB methods separately.}
}
\label{fig:visDA}
\vspace{-0.1in}
\end{figure}

\new{
Our initial trial using adversarial training shows improvement when there is no labeled target $L_t=\emptyset$.
However, after having a few labeled targets, using adversarial training does not introduce further improvement (see supplementary material).
We argue that, 1) the domain gap (from synthetic to real images) in this dataset is large, thus the benefit of aligning image features from target to source domain is less than adding annotated target images $L_t$, and 2) due to the use of ImageNet pre-trained model, the target domain (images from MS-COCO~\cite{lin2014microsoft}) is actually closer to the domain for pre-training (images from ImageNet~\cite{russakovsky2015imagenet}) than the source domain (synthetic images).
}
%

Based on the above observations, we use fine-tuning as our training scheme on VisDA-18, and compare different sampling strategies in Figure~\ref{fig:visDA}.
We set $b=100$ and perform 20 rounds in total.
Using importance weight for sampling performs on a par with BvSB and K-center, and outperforms K-means and random selection baselines.
%
%

%


\section{Conclusion}
\new{
We propose AADA, a unified framework for domain adaptation and active learning via adversarial training. 
When few labeled targets are available, the domain adversarial model helps improve the classification; meanwhile, the discriminator can be utilized to obtain the importance weight for active sample selection in the target domain.
We conduct extensive ablation studies and analyses, and show improvements over other baselines with different training and sampling schemes on object recognition and detection tasks.
In the future, we will consider extending our work to other settings such as open set \cite{Saito_openset_eccv18}, partial \cite{Cao_partialda_eccv18}, and universal \cite{You_universalda_cvpr19} domain adaptation. 
} 

\vspace{-0.1in}
{\small{\paragraph{Acknowledgements} The project is supported in part by NSF IIS \#1749833.}}

{\small
\bibliographystyle{ieee}
\bibliography{egbib}
}

\clearpage
\appendix
\section{Supplementary Material}
In this supplementary material, we include: (1) comparison between AADA and ALDA~\cite{saha2011active} on digit classification, (2) performance on object detection after more selection rounds, (3) comparison of different training schemes on the Office dataset~\cite{office}, (4) comparison of adversarial training and fine-tuning on the VisDA dataset~\cite{peng2018visda,peng2018syn2real}, (5) effect of source data number, and (6) results using CDAN~\cite{long2018conditional}.

\subsection{Comparison to ALDA~\cite{saha2011active}}
\new{Although joint training with importance sampling is one way to extend the ALDA~\cite{saha2011active} method (as compared in Section~4.3 in the main paper), here we consider the original algorithm of online ALDA (O-ALDA) on digit classification. We first extract the features from our domain adversarial model and train a perceptron classifier $u_\phi$, a source classifier $w_{src}$, and a domain separator $w_{ds}$ separately. There are two main differences: 1) this algorithm is performed in an online version, \ie, selecting one sample at a time then updating the classifier, and 2) if the selected image is similar to the source domain (determined by $w_{ds}$), we use the pseudo-label from $w_{src}$ without cost, and hence the number of selected images maybe be larger than the actual budget. The results are shown in Table~\ref{tab:ALDA}, and our method outperforms O-ALDA by 10-15\%.
}
\begin{table}[ht!]
\small{
	\centering
	\renewcommand{\arraystretch}{1}
    \setlength{\tabcolsep}{6pt}
	\begin{tabular}{c|c|c|c|c|c|c}
		\toprule
		\multirow{2}{*}{Method} & \multicolumn{6}{c}{Number of Labeled Target} \\
		\cline{2-7}
		 & 0 & 100 & 200 & 300 & 500 & 1000\\
		\hline
		AADA (Ours)        & 76.5 & \textbf{94.1} & \textbf{95.1} & \textbf{95.6} & \textbf{96.9} & \textbf{97.5}\\
		O-ALDA~\cite{saha2011active} & 76.5 & 79.0 & 81.4 & 82.7 & 84.1 & 87.7\\
		\bottomrule
	\end{tabular}
	\vspace{1.0mm}
	\caption{
	\new{Comparison of AADA and O-ALDA~\cite{saha2011active} on digit classification (SVHN $\rightarrow$ MNIST).}
	}
	\label{tab:ALDA}
}
\end{table}

\subsection{More Object Detection Results}
\begin{figure}[t!]
\centering
    \includegraphics[clip, width=0.99\linewidth]{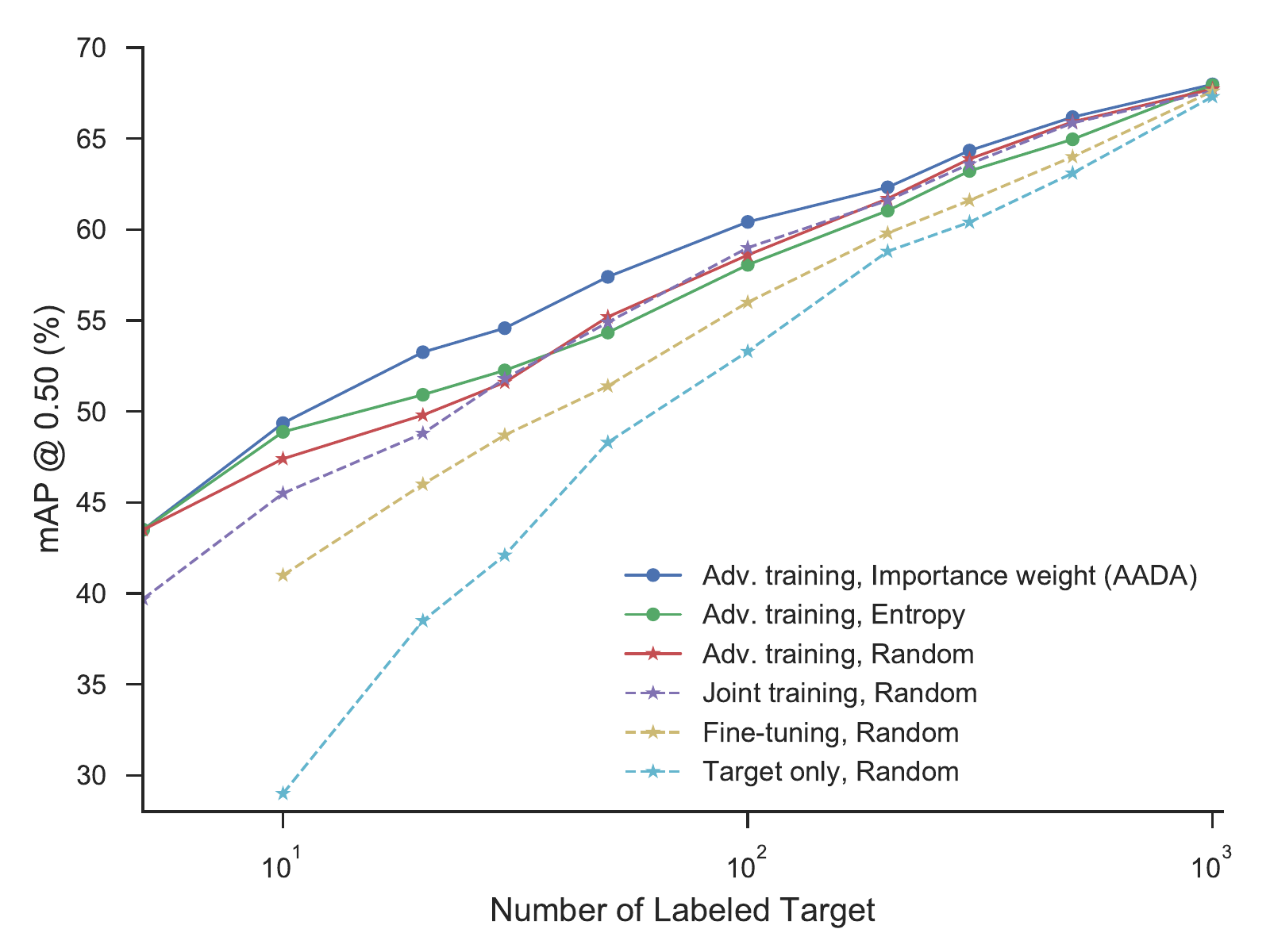}
\caption{\new{Object detection result (KITTI $\rightarrow$ Cityscapes) after 9 rounds. The $x$-axis is shown in log scale. The left-most points represent the initial round where no labeled target is available. Our AADA outperforms all other baselines when up to 1000 labeled targets are available.}}

\label{fig:detection_more}
\end{figure}

Here we show the results on object detection after more sample selection rounds, which is an extension of Table~1 in the main paper. 
We perform 9 rounds in total with $b=$\{10, 10, 10, 20, 50, 100, 100, 200, 500\} for each round. 
We plot the $x$-axis in log scale for a better illustration in Figure~\ref{fig:detection_more}.
Our AADA improves over other baselines, including other sampling strategies with adversarial training and random sampling with different training schemes when up to 1000 labeled targets are available.
%
%

\subsection{Comparison of Training Schemes on Office}
\begin{figure}[t!]
\centering
    \includegraphics[clip, width=0.99\linewidth]{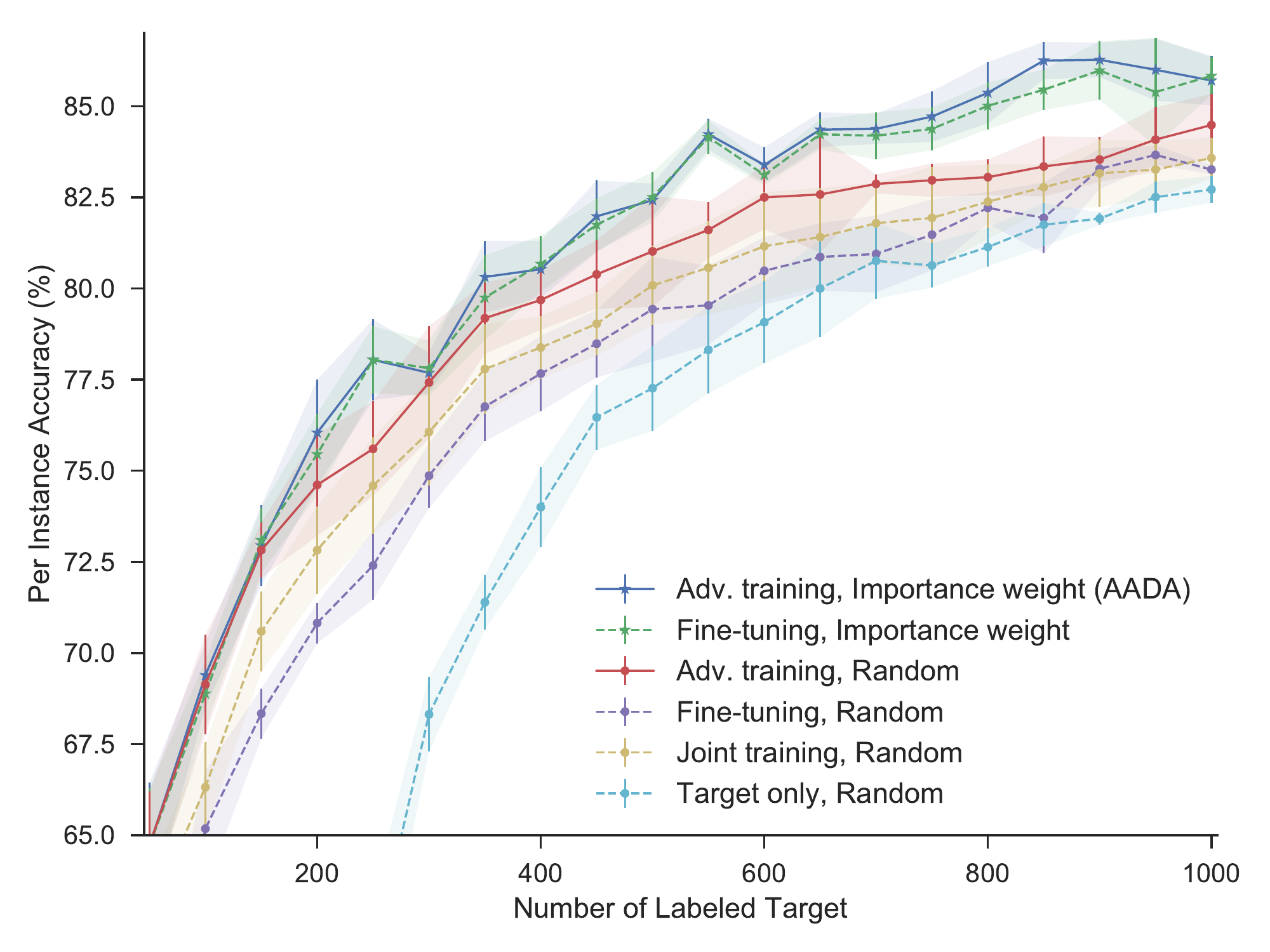}
\caption{Comparing different training schemes on the Office dataset (D $\rightarrow$ A). Adversarial training with importance weight for sampling (AADA) outperforms other baselines with different training schemes.}
\vspace{-0.1in}
\label{fig:office_ablation}
\end{figure}

In this section, we compare the results of adversarial training with different training schemes on the Office dataset~\cite{office} in Figure~\ref{fig:office_ablation}, as an extension of Section~5.1 in the main paper.
With a random selection, adversarial training is better than other baselines including fine-tuning, joint training, and train on target data only. 
When using importance weight for sampling, adversarial training outperforms fine-tuning baseline.
In addition, sampling with the proposed importance weight improves the performance over random selection when either adversarial training or fine-tuning is used.
Overall, our adversarial training with importance weight (AADA) performs the best compared to other combinations of training schemes and sampling strategies.
%

\subsection{Comparison of Training Schemes on VisDA}
\begin{figure}[t!]
\centering
    \includegraphics[clip, width=0.99\linewidth]{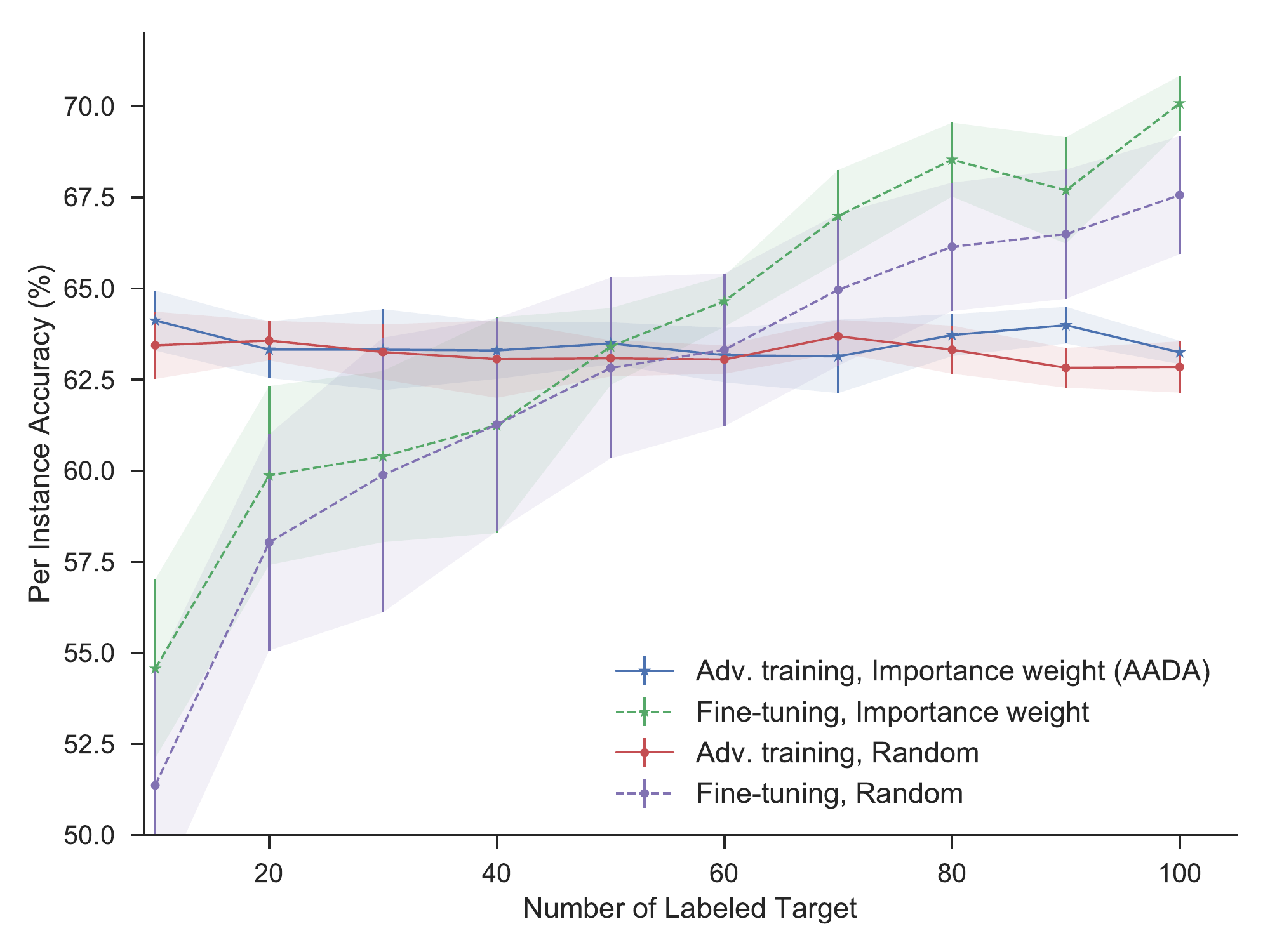}
\caption{Comparing different training schemes on the VisDA dataset. Using adversarial training, the accuracy does not improve when more labeled targets are added since the target domain in VisDA is closer to ImageNet images for pre-training. However, the accuracy improves when we use fine-tuning, in which using importance weight for sampling is better than random sampling.
}
\label{fig:visda_ablation}
\end{figure}

As described in Section~5.3 in the main paper, VisDA~\cite{peng2018visda,peng2018syn2real} is a special case where the target domain is closer to images from ImageNet which is used for pre-training, and thus we utilize the fine-tuning strategy.
In Figure~\ref{fig:visda_ablation}, we further provide results of using adversarial training when few labeled targets $L_t$ are available.
%
To show more fine-grained results, we sample 10 images per round, \ie, $b=10$, and perform 10 rounds of selection.
In an unsupervised domain adaptation setting, \ie, no labeled target is available $L_t=\emptyset$, using adversarial training on $(L_s,U_t)$ improves the test accuracy on the target domain from 57.0\% to 62.5\%, compared to the model trained only on labeled source $L_s$ \emph{without} adaptation.
However, after adding labeled targets, the accuracy of the model using adversarial training decreases, as shown in blue and red curves in Figure~\ref{fig:visda_ablation}, regardless of which sampling strategy is used.
On the other hand, the accuracy of the model using fine-tuning increases when the number of labeled target increases, showing that VisDA is more suitable for fine-tuning due to its dataset property.
Nevertheless, fine-tuning with our proposed importance weight still performs better than random sampling.

\subsection{Effect of Source Data Number}
\begin{figure}[ht!]
\centering
    \includegraphics[clip, width=0.99\linewidth]{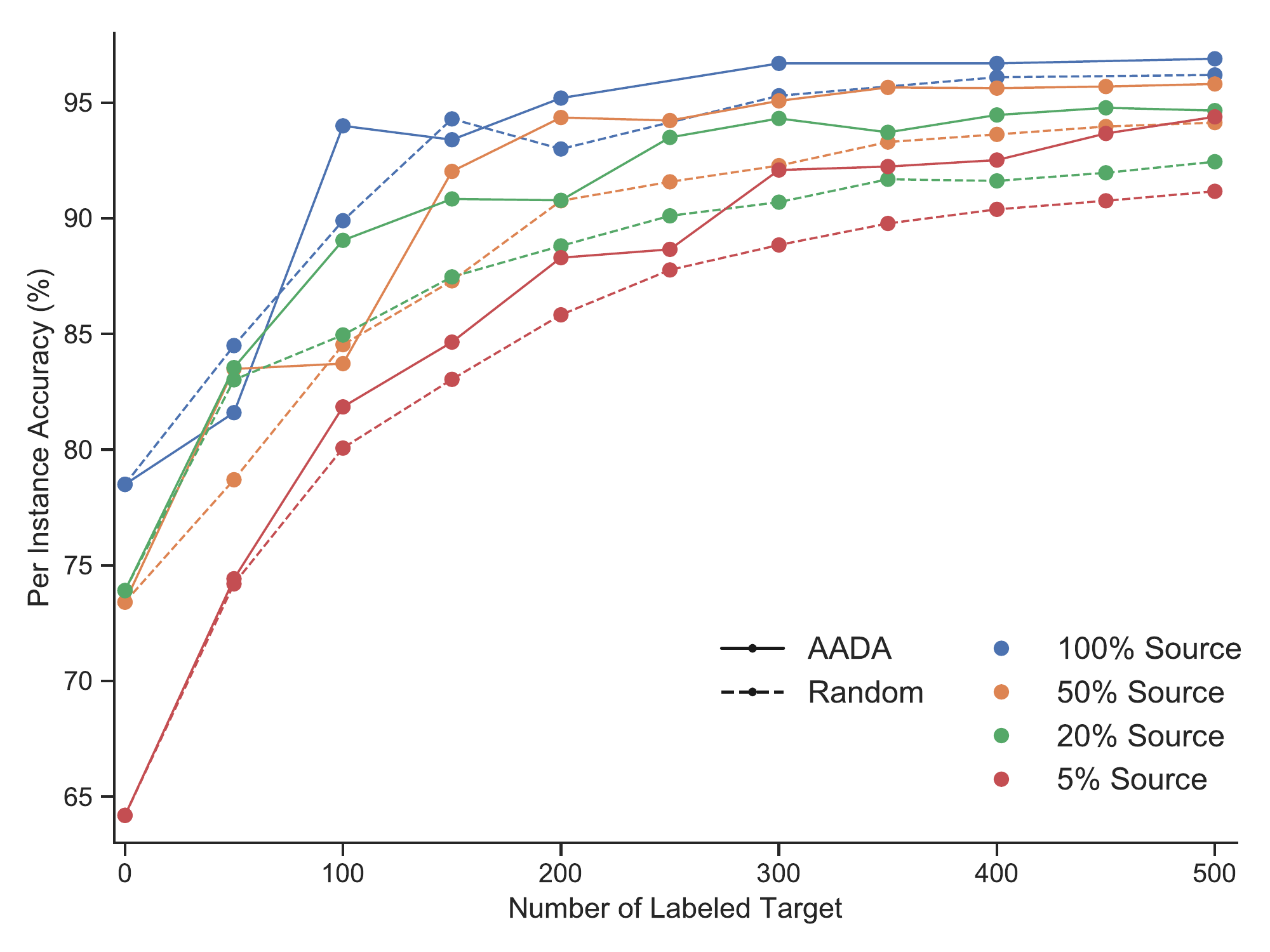}
\caption{Results on SVHN $\rightarrow$ MNIST with a subset of labeled source data. Using importance weight for sampling is better than random selection, and the improvement is even higher when we have less labeled source data for training.}
\label{fig:vary}
\end{figure}

We investigate the effectiveness of our method when the labeled source data is also limited. We use a subset \{5,20,50\}\% of the source data, and compare results using adversarial training. We select 50 labeled targets per round and perform 10 rounds in total. As shown in Figure~\ref{fig:vary}, using importance weight improves over random sampling in all the cases, especially on a smaller subset.

\subsection{Results on Digits with CDAN~\cite{long2018conditional}}
Our AADA framework can be applied to any domain adaptation model with a domain classifier and adversarial training. Here we integrate CDAN~\cite{long2018conditional} model in our AADA framework and experiment on digit classification (SVHN$\rightarrow$MNIST). We use the implementation provided by the authors and follow their training procedure, which yields 87.8\% accuracy when there is no labeled target.
We select 10 labeled targets in each round, and the performance saturates after 50 rounds. We compare different sampling methods in Table~\ref{tab:CDAN}. Our proposed importance weight performs favorably against other methods. 
\begin{table}[t!]
\footnotesize{
	\renewcommand{\arraystretch}{1}
    \setlength{\tabcolsep}{8pt}
    \centering
    \begin{tabular}{c|c|c|c|c|c|c}
		\toprule
		\multirow{2}{*}{Sampling} 
		& \multicolumn{6}{c}{Number of Rounds} \\
		\cline{2-7}
		 & 5 & 10 & 15 & 20 & 30 & 50 \\
		\hline
		Imp.~weight & \textbf{92.1} & \textbf{94.2} & 94.7 & \textbf{95.2} & \textbf{95.5} & 95.8 \\
        BvSB & 92.0 & \textbf{94.2} & \textbf{94.8} & 95.0 & 95.2 & \textbf{95.9} \\
        Entropy & 91.1 & 93.1 & 94.5 & 94.6 & 95.0 & 95.6 \\
		Random & 89.9 & 92.9 & 94.3 & 94.7 &  95.2 & 95.5 \\
		\bottomrule
	\end{tabular}
	\vspace{0.05in}
	\caption{
	Results on SVHN $\rightarrow$ MNIST. We use CDAN~\cite{long2018conditional} for training and compare different sampling approaches. In each round, we select 10 target samples to label.
	}
	\label{tab:CDAN}
}
\end{table}

\end{document}